%% file: example_paper.tex
\documentclass{article}
\usepackage{microtype}
\usepackage{graphicx}
\usepackage{subfigure}
\usepackage{booktabs}
\usepackage{hyperref}

\usepackage{bigai2023}
\usepackage{amsmath}
\usepackage{amssymb}
\usepackage{mathtools}
\usepackage{amsthm}
\usepackage[capitalize,noabbrev]{cleveref}

\theoremstyle{plain}

\theoremstyle{definition}

\theoremstyle{remark}

\bigaititlerunning{Short Title for Header}
\begin{document}
\twocolumn[

\bigaititle{S$^{3}$IT: A Benchmark for \underline{S}patially \underline{S}ituated \underline{S}ocial \underline{I}ntelligence \underline{T}est}

\begin{bigaiauthorlist}
Zhe Sun, Xueyuan Yang, Yujie Lu, Zhenliang Zhang $^\dagger$
\end{bigaiauthorlist}
\newline 

State Key Laboratory of General Artificial Intelligence, Beijing Institute for General Artificial Intelligence (BIGAI), Beijing 100080, China

\vskip 0.3in

\bigaikeywords{Situated Social Intelligence, Embodied social intelligence, Embodied Agent, AI benchmark}
\vskip 0.4in
]
\printAffiliationsAndNotice{\textit{  \textsuperscript{$\dagger$} Corresponding author(s): zlzhang@bigai.ai}}

\input{S3IT/0_abstract}    
\input{S3IT/1_Introduction}
\input{S3IT/2_Relatedworks}

\input{S3IT/3_Method}

\input{S3IT/4_Implement}
\input{S3IT/5_Experiment}

\input{S3IT/6_Discussion}
\input{S3IT/7_Conclusion}

{\small
\bibliographystyle{unsrtnat}

\bibliography{example_paper}
}
\end{document}

%% file: S3IT/0_abstract.tex
\begin{abstract}

The integration of embodied agents into human environments demands embodied social intelligence: reasoning over both social norms and physical constraints. However, existing evaluations fail to address this integration, as they are limited to either disembodied social reasoning (e.g., in text) or socially-agnostic physical tasks.
Both approaches fail to assess an agent's ability to integrate and trade off both physical and social constraints within a realistic, embodied context. 
To address this challenge, we introduce Spatially Situated Social Intelligence Test (S$^{3}$IT)
, a benchmark specifically designed to evaluate embodied social intelligence. It is centered on a novel and challenging seat-ordering task, requiring an agent to arrange seating in a 3D environment for a group of large language model-driven (LLM-driven) NPCs with diverse identities, preferences, and intricate interpersonal relationships. Our procedurally extensible framework generates a vast and diverse scenario space with controllable difficulty, compelling the agent to acquire preferences through active dialogue, perceive the environment via autonomous exploration, and perform multi-objective optimization within a complex constraint network. 
We evaluate state-of-the-art LLMs on S$^{3}$IT and found that they still struggle with this problem, showing an obvious gap compared with the human baseline. Results imply that LLMs have deficiencies in spatial intelligence, yet simultaneously demonstrate their ability to achieve near human-level competence in resolving conflicts that possess explicit textual cues.

\end{abstract}

%% file: S3IT/1_Introduction.tex
\section{Introduction}
\begin{figure}[t]
    \centering
    \includegraphics[width=\linewidth]{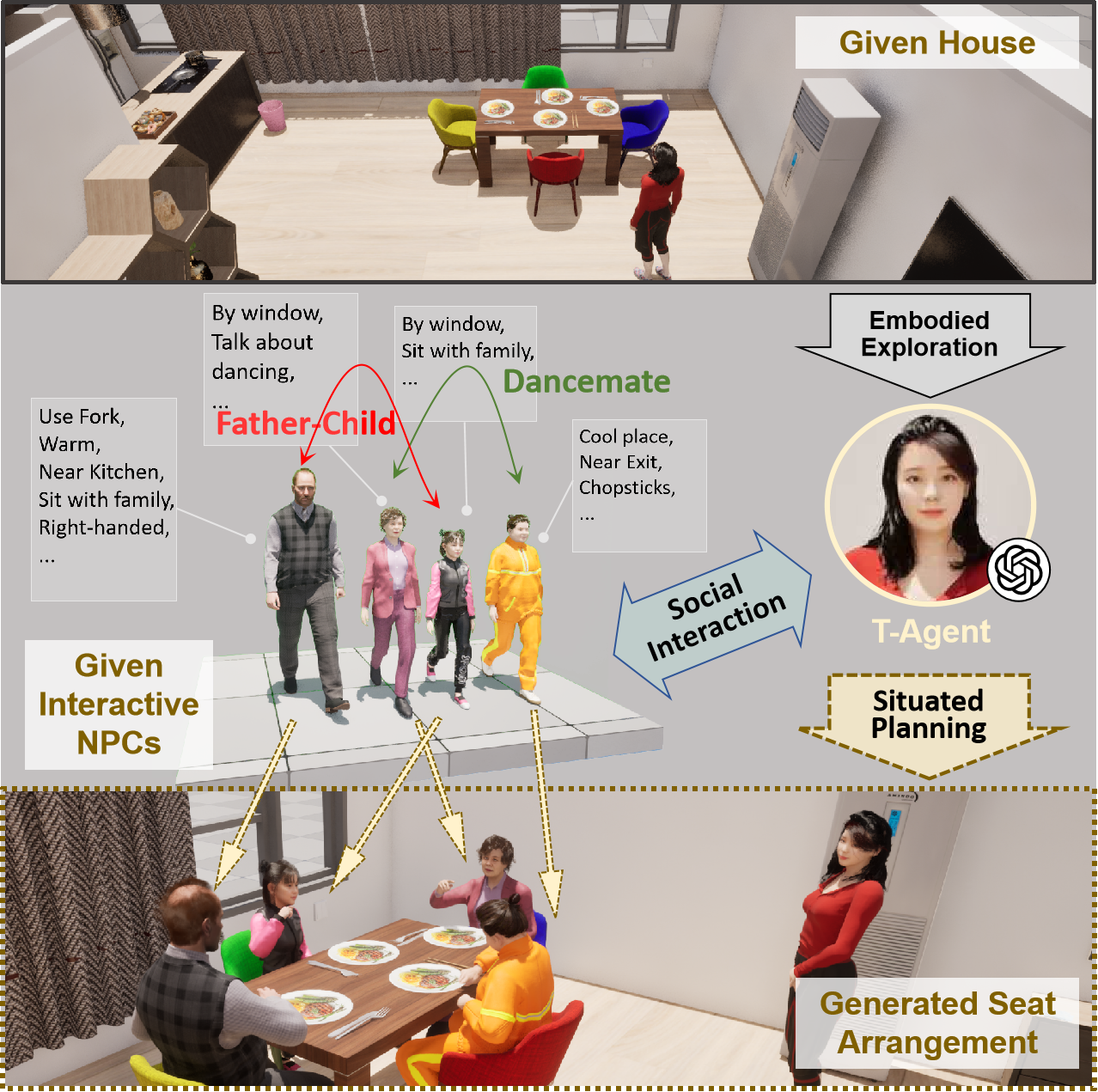}\vspace{-10pt}
    \caption{A typical seat arrangement task involves a given room layout and several NPCs. The agent under test (T-Agent) needs to interact with the NPCs and explore the room to devise a seating arrangement that satisfies everyone.}
    \label{fig:teaser}
\end{figure}
As agents driven by large language models (LLMs) become increasingly integrated into human environments, they are evolving from digital entities into embodied collaborators. To achieve effective and reliable human-agent collaboration, these agents need abilities not only to execute physical tasks but also to understand and adhere to complex social norms, which is known as ``social intelligence” \cite{Kihlstrom2000SocialIntelligence, smith2007situated, zhouvirtual,sternberg2000handbook}. A foundational challenge in developing such intelligence is enabling agents to reason about explicit social rules, preferences, and relationships, especially when this reasoning must occur within physically grounded contexts. This requires agents to seamlessly integrate social understanding with spatial awareness, a capability that is critical for their successful deployment in real-world human spaces yet remains underexplored.

The path toward social intelligence is characterized by reducing the levels of abstraction that separate agents' training environment from reality \cite{zhang2024emergence}. Early benchmarks were highly abstract, confining evaluations to static narratives \cite{nematzadeh2018evaluating,le2019revisiting,sap2022neural, shapira2023clever}. 
This allowed models to potentially solve tasks through heuristic shortcuts or by exploiting spurious correlations, rather than by achieving true reasoning skills\cite{sclar2023minding,shapira2023clever,ullman2023large}.
The introduction of interactive dialogue \cite{chan2024negotiationtom}
represented a significant reduction in abstraction, yet this approach fundamentally treated LLMs as parsers of linguistic symbols rather than as actors capable of interacting with an environment. More recently, researchers have moved closer to realism by endowing agents with proactive agency in tasks like multi-agent games \cite{agashe-etal-2025-llm}.
Nevertheless, these simplified and abstracted environments remain a high-level abstraction from, rather than a faithful representation of, true physical environment.

Concurrently, several studies have explored human-agent collaboration within three-dimensional (3D) embodied environments, demonstrating that agents possess a certain degree of social awareness \cite{puig2023habitat,du2024constrained}. 
However, their predominant focus has been on the execution of physical tasks (e.g., fetching an object), while complex social reasoning has often been secondary. The academic community has yet to reach a consensus on an evaluation paradigm for such social capabilities. The question of how to systematically evaluate an embodied agent's ability to understand and adhere to complex social norms remains open.

We posit that the evaluation of an agent's social intelligence must be rooted in interactive, contextualized, and embodied scenarios.
To this end, we address the challenge of \textbf{Spatially Situated Social Intelligence Test (S$^{3}$IT)} and introduce a novel benchmark for its systematic evaluation. An example is shown in Figure~\ref{fig:teaser}.
The core task of this benchmark is to enable an embodied agent to interact with a group of non-player characters (NPCs), drawn from a persistent world of 59 residents, in a simulated 3D environment and assign them seating arrangements. We denote the agent under test as \textbf{T-Agent}. These NPCs possess diverse preferences and intricate relationships. The framework is highly modular and scalable, allowing for programmatic generation of a vast and adjustable task space by varying parameters such as room layout, NPC preferences, and interpersonal conflicts. 
This design compels the T-Agent to move beyond simple instruction-following and instead proactively discover, comprehend, and weigh a complex network of constraints interwoven from both physical and social constraints, which encourages agents to arrive at the most appropriate decision based on the given information.

Our benchmark includes a three-stage evaluation pipeline: 1) NPC Preference Extraction and Summarization: The T-Agent interacts with NPCs to extract and summarize their preferences, constructing a comprehensive ``preference profile''; 2) Environment Cognition: The T-Agent proactively explores the 3D scene to build a cognitive map of the space; 3) Multi-Constraint Decision-Making: Integrating information from the previous two stages, the T-Agent generates and iteratively refines a seating arrangement. This process comprehensively evaluates the T-Agent's integrated ability to synthesize multimodal information, perform spatial reasoning, and engage in advanced social inference.

We also introduce an automatic evaluation pipeline and benchmark several state-of-the-art (SOTA) LLMs. The findings demonstrate that these models, while proficient at text-based information extraction, struggle to handle constraints involving spatial and social constraints, revealing a significant deficit in their holistic embodied social intelligence.

The main contributions of this paper are:
\begin{itemize}

    \item We introduce S$^{3}$IT, a novel benchmark to evaluate an embodied agent's social reasoning and planning under intertwined physical and social constraints.
    
    \item We design a procedurally extensible framework for task generation and evaluation. This framework can systematically construct test scenarios of controllable difficulty and diverse types, enabling a fine-grained analysis of an agent's capabilities and limitations.
    
    \item We systematically evaluate leading LLMs on embodied social reasoning, highlighting their capabilities in conflict handling and limitations in spatial intelligence with embodied and social constraints.
 
\end{itemize}

%% file: S3IT/2_Relatedworks.tex
\section{Related Works}
\subsection{Social Intelligence}
Social intelligence refers to the broad capability of an agent to navigate social environments, understand social cues, and interact appropriately with other agents \cite{Kihlstrom2000SocialIntelligence, smith2007situated, zhouvirtual,sternberg2000handbook}. Within this field, research has explored various facets of cognition. A significant line of inquiry focuses on Theory of Mind (ToM)—the ability to attribute unobservable mental states like beliefs and intentions to others \cite{frith2005theory}. In contrast, our work addresses a distinct yet equally critical challenge: embodied social reasoning. This capability pertains to reasoning about explicit social information, such as stated preferences and interpersonal relationships, and integrating it with constraints from the physical world. It focuses on how agents can act effectively in socially and spatially complex situations based on available information.

\subsection{Narrative-Based Social Reasoning Evaluation} 
Current evaluation paradigms for social reasoning in LLMs predominantly rely on narrative-based benchmarks \cite{nematzadeh2018evaluating,le2019revisiting,sap2022neural, shapira2023clever,gandhi2023understanding,he2023hi,gu2024simpletom,strachan2024testing}, including some with multimodal extensions~\cite{shi2025muma,jin2024mmtom}. The majority of these benchmarks probe a model's performance on specific cognitive dimensions, particularly belief reasoning, by adapting classic ``false-belief'' tasks such as the Sally-Anne test~\cite{baron1985does}. However, such evaluation frameworks possess inherent limitations: they tend to position the LLM as a passive observer analyzing static text. This paradigm not only stands in stark contrast to how humans interact within dynamic and continuous real-world environments but may also enable models to ``solve'' tasks by leveraging spurious statistical cues rather than acquiring genuine reasoning abilities~\cite{sclar2023minding,shapira2023clever,ullman2023large,ma2023towards}. Although recent, more complex benchmarks~\cite{kim2023fantom,yu2025persuasivetom,chan2024negotiationtom} have introduced interactive dialogues with information asymmetry to mitigate these issues, their core framework still treats the LLM as an analyzer of linguistic symbols, rather than as an agent situated within an environment.

\subsection{Interactive Paradigms for Social Reasoning Evaluation}To overcome these limitations, a research paradigm has emerged, which evaluates the social capabilities of LLMs by embedding them as agents in goal-oriented, multi-agent tasks. Research has demonstrated that explicit mental reasoning enhances performance in multi-agent coordination \cite{gu2024simpletom,lim2020improving}. This thesis has been validated in environments such as Overcooked-AI, Hanabi, and various text-based collaborative games like the ``Guandan'' card game \cite{agashe-etal-2025-llm,li-etal-2023-theory,guandan}. In these settings, agents must act as participants, formulating strategies by inferring the intentions and beliefs of their partners or adversaries. 
While these methods mark a critical step toward practical social intelligence by situating reasoning in active, interactive contexts, they mostly operate within highly simplified game or sandbox environments. Consequently, they largely decouple social reasoning from the rich spatial and physical reasoning inherent to embodied interaction. Effectively integrating these two abilities remains a frontier challenge in the field of social intelligence.

\subsection{Embodied Social Reasoning}Building on seminal calls to advance social intelligence evaluation \cite{ma2023towards,wang2025rethinking}, we argue that robust assessment requires grounding agents in embodied, interactive environments. Recent research \cite{puig2020watch,puig2023habitat,du2024constrained} represents significant strides in this direction. However, their core focus remains on physical collaboration, where agents infer a human's physical goals from visual observation to assist in task execution. This paradigm, however, fails to encompass the richer and more nuanced social dimensions fundamental to human interaction. To address this limitation, we propose shifting from purely physical task-solving to embodied social reasoning, wherein agents must navigate complex physical and social constraints.

%% file: S3IT/3_Method.tex
\section{Method}
We propose a benchmark for situated social intelligence in embodied agents with the following features:
\begin{enumerate}
    \item 3D Simulated Environment: The T-Agent is placed in an explorable 3D space, allowing the evaluation of its spatial exploration and understanding abilities.
    \item Social Interaction with NPCs: Several embodied NPCs with a certain degree of automation are present. The T-Agent can interact with them, allowing the evaluation of its social skills and ability to acquire new information through communication.
    \item Complex Constraints: Both spatial and social constraints are imposed to test the T-Agent's ability to balance and plan in complicated social situations, as well as its understanding of NPCs' preferences.
\end{enumerate}
The seat-ordering problem meets all these requirements. It is clear, understandable for both agents and peers, and NP-hard, providing sufficient challenges. Thus, we selected it as the baseline task.

\subsection{Worldview and Background Generation}
We designed a small town of 59 residents. Each resident has their name, unique 3D appearance, age, gender, job, workplace, residence, income level, and social relationships.
We also set their interests and dominant hand, as well as three kinds of preferences: 1) embodied preference, 2) social preference, and 3) social conflict.

The 59 residents in our dataset are organized into 11 families, spanning up to 4 generations. We designed a wide variety of family structures, including nuclear, single-parent, reconstituted, only-child, and multi-child (e.g., with twins) families. These give rise to primary kinship ties (spouse, parent, child, sibling), extended ties (grandparent, grandchild), and affinal ties (in-laws). The dataset also incorporates specific child-rearing contexts, such as families with left-behind children. Beyond kinship, we defined social relationships based on geographical proximity (neighbors, friends) and division of labor (colleagues, superiors, subordinates, teachers, students, classmates). Details are included in the supplementary materials.

\begin{figure}
    \centering
    \includegraphics[width=0.8\linewidth]{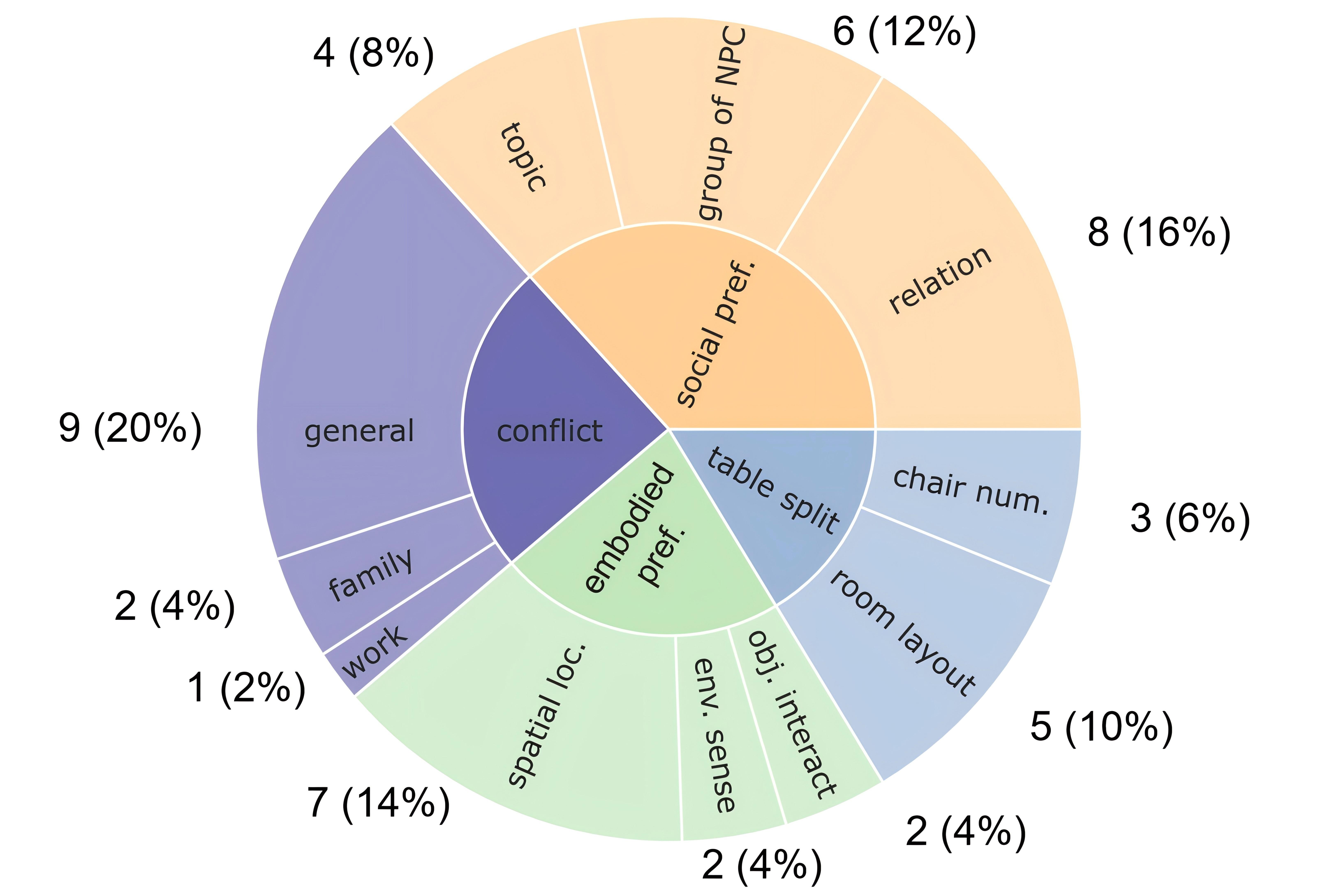}\vspace{-12pt}
    \caption{The four groups of elements: 1) table splitting, 2) embodied spatial understanding (embodied pref.), 3) social relationship understanding (social pref.), and 4) conflicts, as well as the categories they contain.}
    \label{fig:element}
\end{figure}

\subsection{Automatic Question Generation}
To enhance the diversity of the dataset, we proposed 4 groups of elements to be involved in a question: 1) table splitting, 2) embodied spatial understanding, 3) social preference understanding, and 4) conflict avoidance. Figure \ref{fig:element} shows categories included in each group.

\begin{figure}
    \centering
    \includegraphics[width=\linewidth]{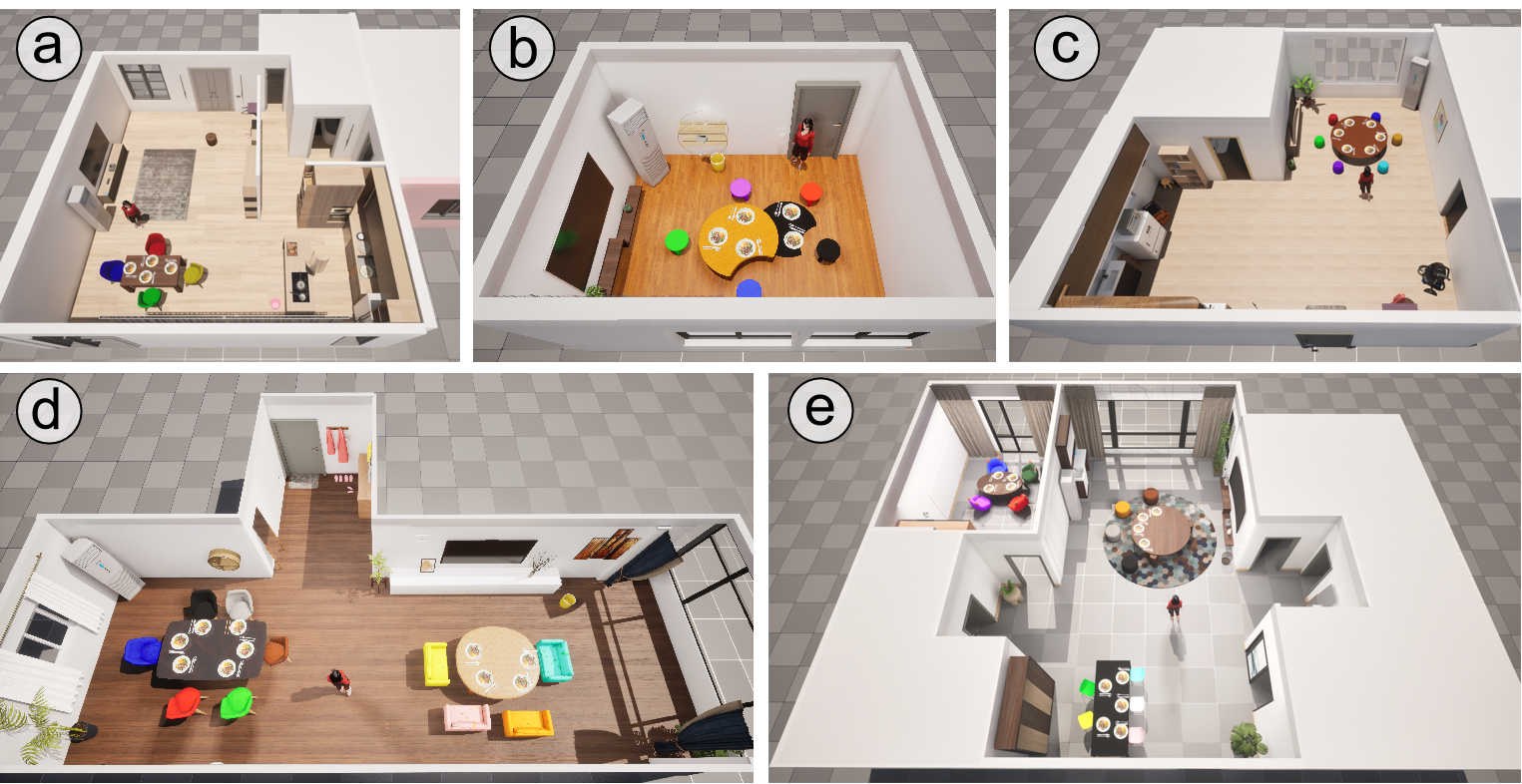}\vspace{-10pt}
    \caption{Five kinds of house layouts. (a) Single rectangular table with 4 chairs. (b) Single irregular table with 5 chairs. (c) Single circular table with 6 chairs. (d) A single room with a rectangular table (6 chairs) and a circular table (4 chairs). (e) Multiple rooms with a rectangular table (5 chairs), a circular table (4 chairs), and an oval table (4 chairs).}
    \label{fig:room}
\end{figure}

\textbf{Table Splitting.}
We proposed a challenging environment based on five distinct scene templates. These templates serve as blueprints, allowing us to generate an expansive, near-infinite set of unique room configurations by randomizing the placement of furniture (tables and chairs). We present five specific, representative room instances derived from these templates in Figure~\ref{fig:room}.

The templates are categorized by complexity, exhibiting variations in table numbers, table shapes, and room numbers. The template set comprises three layouts with a single table, one layout featuring multiple tables within a single room, and one complex layout spanning multiple rooms.

The T-Agent must explore these highly diverse spaces to understand the distribution of tables and chairs and assign NPCs accordingly. To further enhance data diversity, we utilized 8 tables with different shapes and textures and 6 chairs of different shapes during the randomization process.

\begin{figure}
    \centering
    \includegraphics[width=\linewidth]{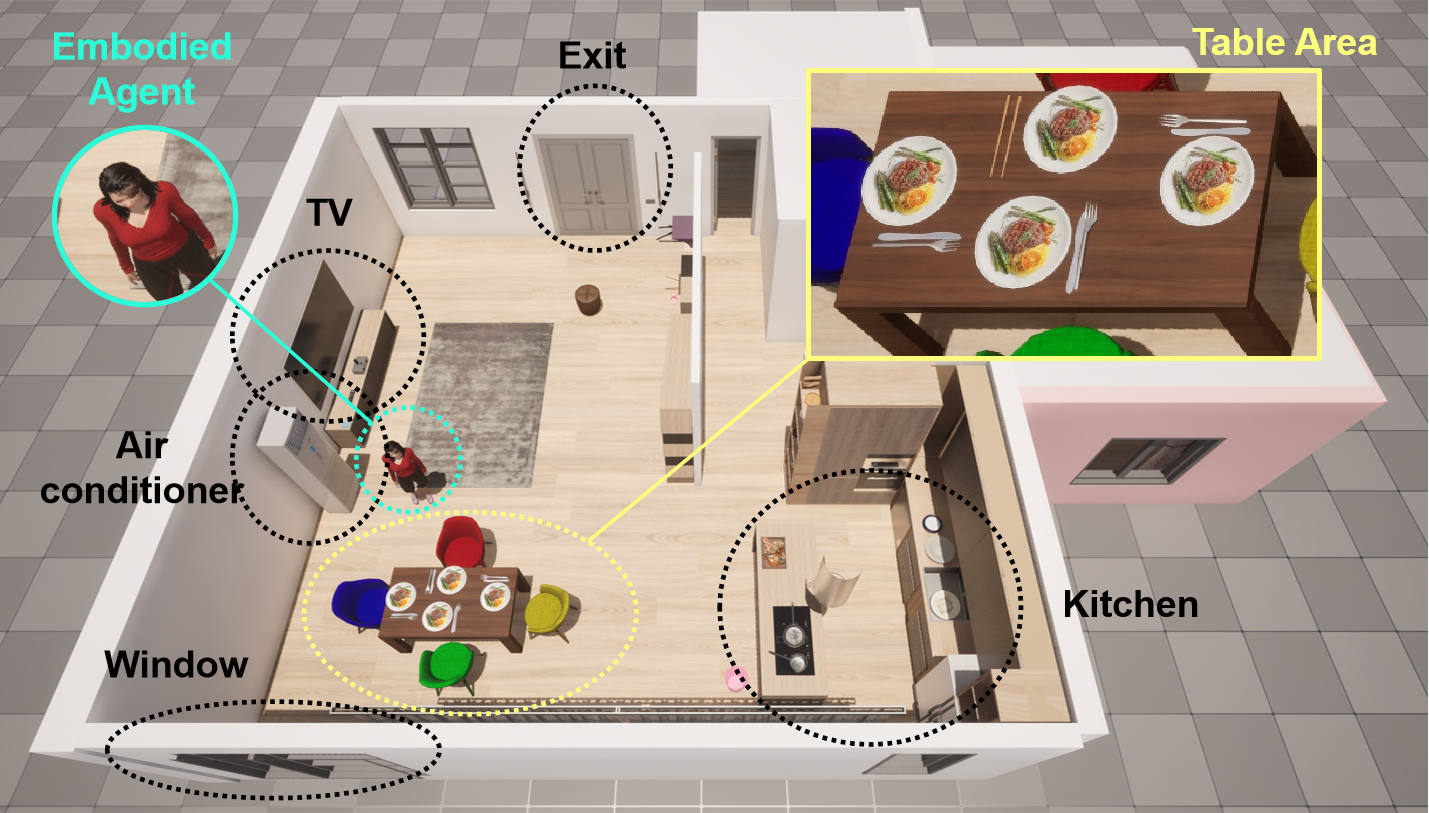}\vspace{-10pt}
    \caption{Detailed preview of House A. The preferences of NPCs originate from the spatial constraints within rooms, as outlined by the black dotted lines in the figure. For instance, the air conditioners relate to their preferences for  temperature, and kitchens correspond to NPCs' tendencies to be near or away from kitchens. Additionally, NPCs have specific requirements for tableware (chopsticks or cutlery), as shown in the enlarged view of the table area.}
    \label{fig:roomdetail}
\end{figure}

\textbf{Embodied Spatial Understanding.}
To measure the situated social intelligence of the T-Agent, it is crucial to evaluate its ability to comprehend the current situation. Hence, we have established rich spatial constraints, endowing each NPC with up to 11 kinds of embodied preferences. The T-Agent explores the house through a given embodiment to grasp spatial features and align them with the NPCs' embodied preferences. Taking one house as an example, Figure~\ref{fig:roomdetail} illustrates the generation of these spatial constraints. Within these houses, we have designated various exits, room layouts, furniture arrangements, table positions, and tableware positions. The embodied preferences of NPCs are derived from a semantic abstraction of the spatial relationship between a chair and the aforementioned spatial constraints, as shown in Figure~\ref{fig:element}. Detailed preference items are provided in the supplementary materials.

\textbf{Social Relationship Understanding.}
In addition to embodied preferences, we have assigned 18 kinds of social preferences to the NPCs. The first category relates to the NPC's interpersonal relationships. For example, some NPCs prefer to sit with classmates or family members. The second category relates to specific groups of people, such as sitting with peers. The last one concerns the topics that the NPC is interested in, such as discussing research challenges. Figure~\ref{fig:element} presents the details of these three categories.

\textbf{Conflicts.}
To enhance the realism of the task, we have introduced 12 types of conflicts among NPCs. The possible conflicts between two NPCs are closely related to their social relationships. For example, family conflicts arise between parents and children and work-related issues occur among colleagues. Strangers, however, are assumed to have no conflicts due to their lack of acquaintance. Figure~\ref{fig:element} shows the details of these conflicts.

\subsection{S$^3$IT Dataset}

\textbf{Dataset Generation.}

To guarantee the solvability of problem instances, which is crucial for a principled evaluation of our agent, we employ a constructive, reverse-engineering approach for generation. Specifically, we first establish a ground-truth solution by randomly permuting the NPCs to form an initial, valid seating arrangement. Subsequently, based on this arrangement, we programmatically derive the corresponding constraints for each NPC, such as preferences and conflicts related to their seating position and neighbors. This generation-by-construction methodology inherently guarantees that at least one valid solution exists for every problem instance.

\begin{figure*}[htbp]
    \centering
    \includegraphics[width=\textwidth]{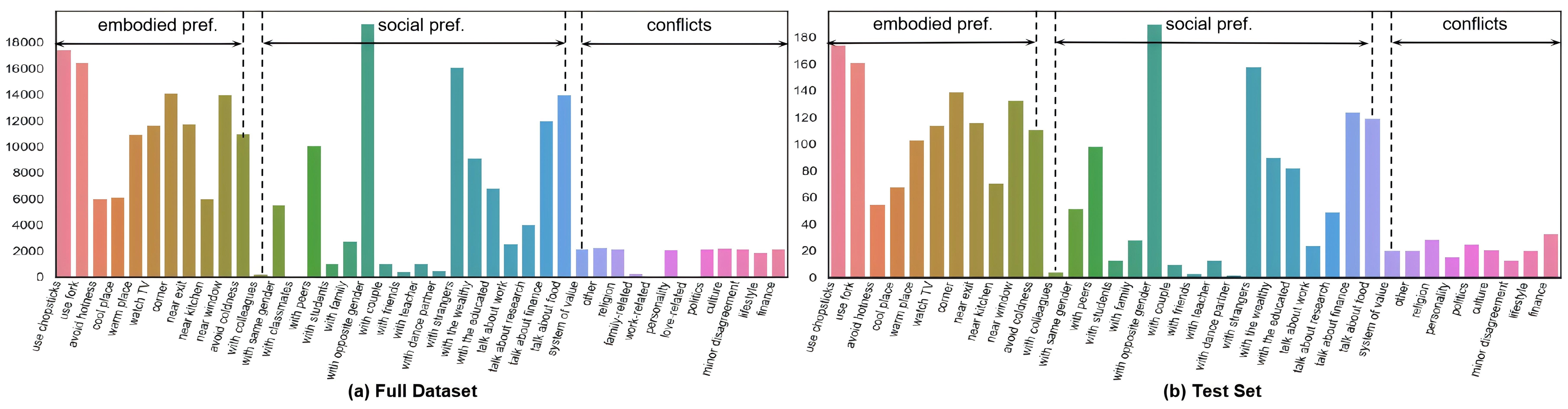}
    \vspace{-24pt}
    \caption{Selection frequency of each preference and conflict. (a) The Full Dataset. (b) The Test Set, whose distribution of elements is consistent with that of the full dataset.}
    \label{fig:dataset}
\end{figure*}

\textbf{Dataset Structure.}
Figure~\ref{fig:dataset}(a) shows the selection frequency of each preference and conflict.
Mindful of the capabilities of current LLM-based agents, we designed a dataset of 7,000 instances across 70 difficulty levels to avoid excessive complexity. These instances are derived from 5 scene templates, where each NPC typically has 1–5 preferences and 0–2 conflicts. 
Each preference and conflict is assigned a strength on a 3-point Likert-type scale. Figure~\ref{fig:dataset}(a) shows the frequency distribution of these generated types.

%% file: S3IT/4_Implement.tex
\section{Testing Pipeline of S$^3$IT Benchmark}
We propose a systematic pipeline for the quantitative evaluation of the social intelligence exhibited by an agent in seating arrangements for diverse social groups. 

The pipeline comprises three phases. In the NPC Preference Extraction and Summarization phase, the T-Agent constructs detailed preference profiles for each NPC. Then, the Environmental Cognition phase tasks the T-Agent with constructing a structured representation of the 3D environment through comprehensive exploration. Finally, during the Multi-Constraint Decision-Making phase, the T-Agent integrates information from the preceding phases to generate, reflect, and iteratively refine seating solutions.

\subsection{Phase I: NPC Preference Extraction and Summarization}
\begin{figure}[tbp]
\renewcommand{\baselinestretch}{1.0}
\centering 
\includegraphics[width=0.95\linewidth, keepaspectratio]{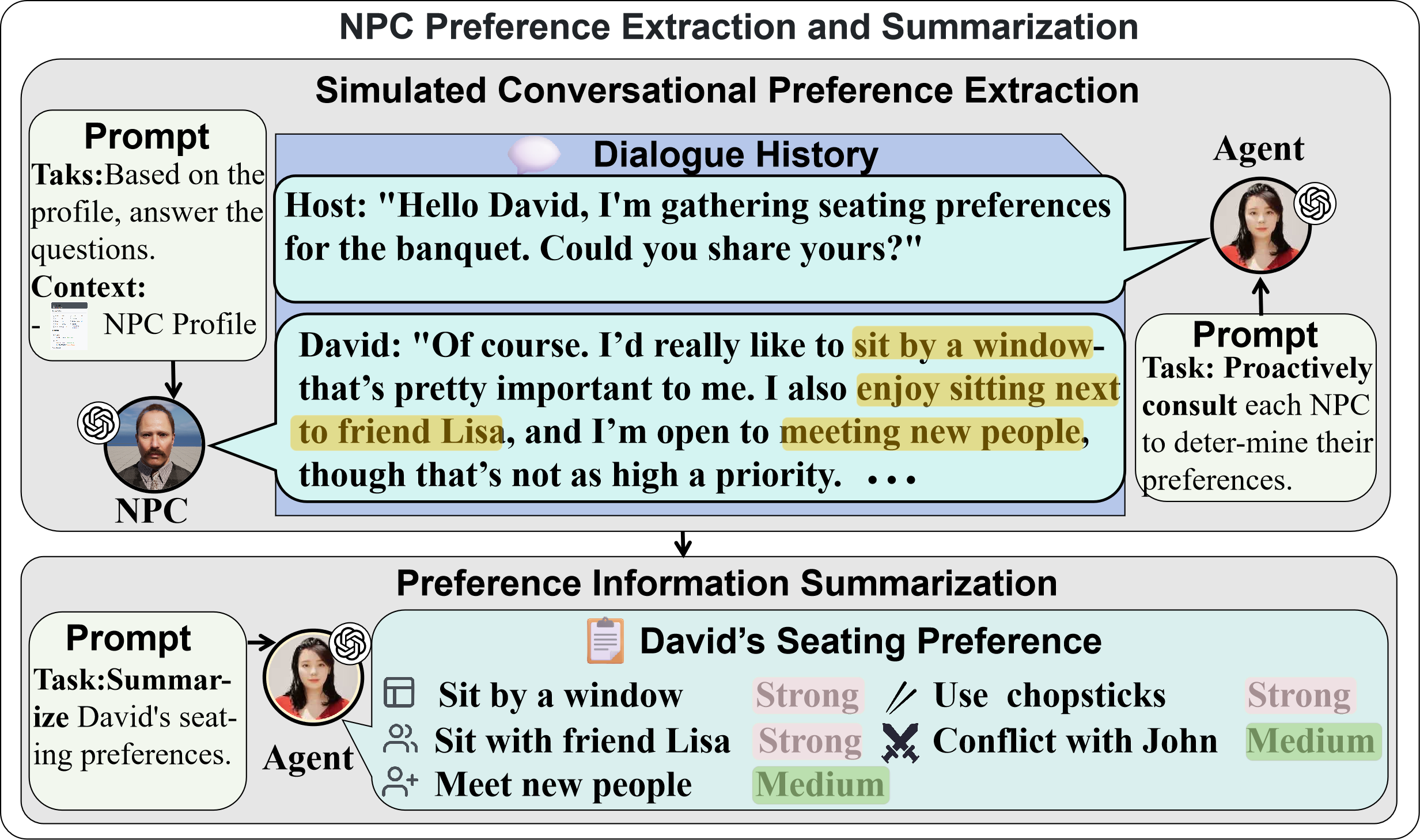} \vspace{-10pt}
\caption{
The T-Agent elicits NPC preferences through simulated conversations, summarizing dialogue into structured profiles.}\label{phase1} \vspace{-12pt}
\end{figure}

To emulate the real-world process of discerning and understanding the needs of others, this phase constructs a structured preference profile for each NPC by extracting and summarizing information from simulated dialogues. These profiles serve as the core constraints for the T-Agent's subsequent decision-making. As illustrated in Figure \ref{phase1}, this process consists of two primary steps:

\textbf{Simulated Conversational Preference Extraction.}
First, each NPC is instantiated from a large language model using a predefined configuration file. The T-Agent then proactively elicits the preferences of NPCs through natural language interaction, with both their basic information (e.g., gender, age) and the complete dialogue history being stored in the memory module.

\textbf{Preference Information Summarization.}
Subsequently, the T-Agent distills the unstructured dialogue into structured NPC preference profiles according to a predefined schema. Each profile includes a description (a concise preference summary) and an intensity score (e.g., strong, medium, or low). These profiles serve as the core constraints for the T-Agent's decision-making and are subsequently stored in the memory module.

\subsection{Phase II: Environmental Cognition}
\begin{figure}[tbp]
\renewcommand{\baselinestretch}{1.0}
\centering 
\includegraphics[width=0.95\linewidth, keepaspectratio]{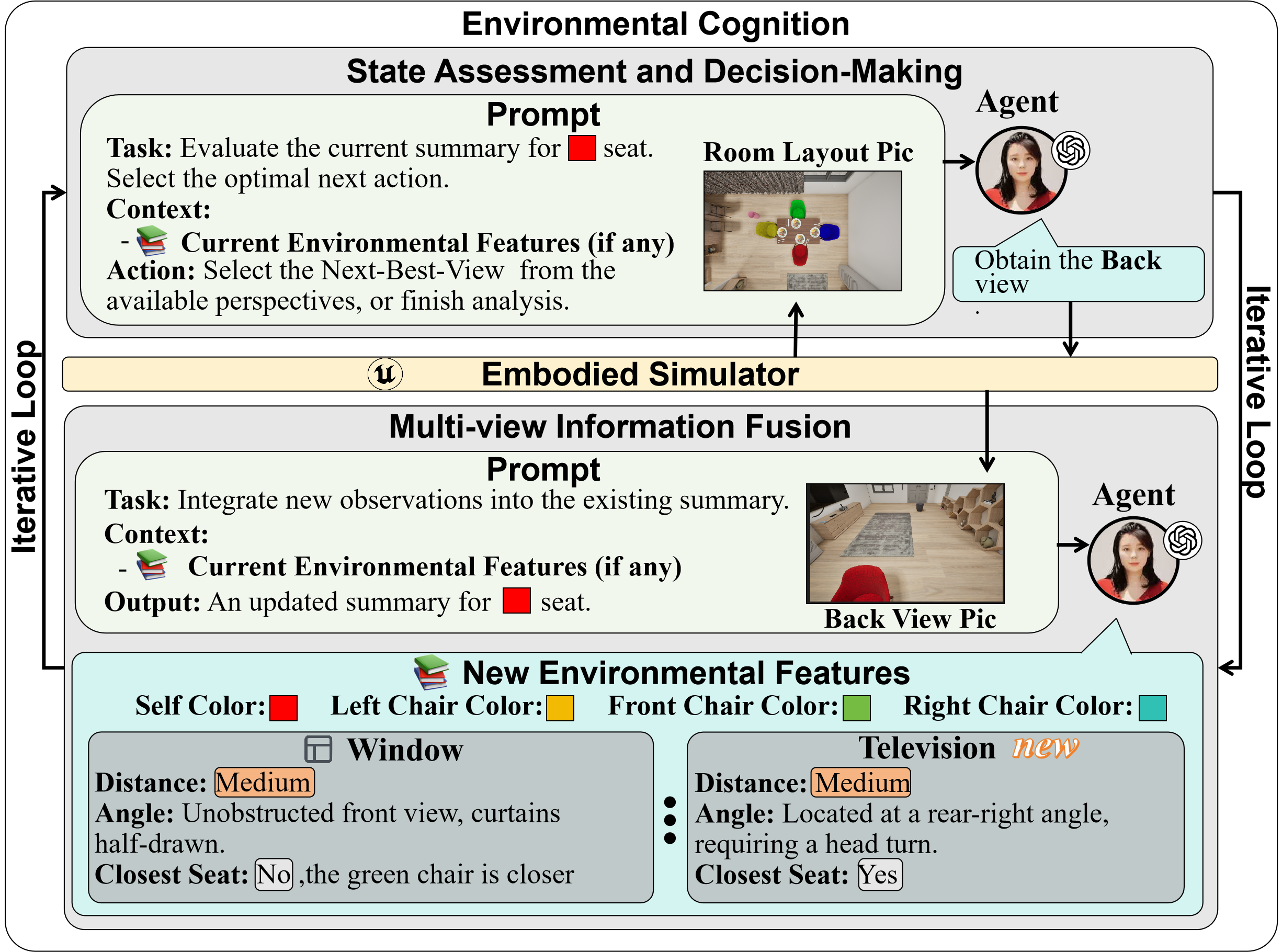}\vspace{-10pt}
\caption{
The T-Agent iteratively constructs environmental features through two stages: 1) State Assessment and Decision-Making, selecting viewpoints to maximize information gain; and 2) Multi-View Fusion, integrating new observations into the feature set. This process repeats until the T-Agent determines that the features are complete or all viewpoints have been explored.}\label{phase2} \vspace{-9pt}
\end{figure}

This phase evaluates the efficacy of the embodied T-Agent in exploring a simulated environment and converting raw visual inputs into structured representations of environmental features. As illustrated in Figure~\ref{phase2}, the T-Agent perceives and interacts with the environment through a target-driven, iterative exploration process (e.g., identifying a ``seat'' and analyzing its surrounding context), which is composed of the following steps:

\textbf{State Assessment and Decision-Making.}
At the start of each iteration, the T-Agent evaluates its current environmental features and the room layout image to inform action selection. If its aggregated understanding remains insufficient, the T-Agent strategically selects a novel, informative viewpoint to maximize information gain. Otherwise, once environmental comprehension is deemed adequate, the T-Agent terminates the exploration process.

\textbf{Multi-view Information Fusion.}
The T-Agent dynamically refines its environmental features by integrating data from novel viewpoints. These features encode not only the properties of seats adjacent to the target, but also quantitatively capture spatial relationships—such as distance, orientation, and viewing angle—between the target and key elements like windows and televisions.

\subsection{Phase III: Multi-Constraint Decision-Making}

This phase is designed to assess the T-Agent's higher-order social reasoning under complex physical and social constraints—including NPCs' profiles and environmental features—in order to resolve the seating allocation problem. As illustrated in Figure \ref{phase3}, this process is implemented as an iterative optimization loop involving ``generation'' and ``reflection'', with the specific steps are as follows:

\textbf{Iterative Solution Generation and Optimization.}
At each iteration, the T-Agent queries the memory module to retrieve NPC profiles and environmental features (serving as static context) and synthesizes a comprehensive seating allocation plan accompanied by a detailed reflection report that explicitly annotates the satisfaction status of each preference. The solution and its corresponding reflection report from the previous iteration are then incorporated as dynamic context for the subsequent round, enabling targeted rectification of unmet preferences. This unified  framework is iteratively executed, progressively refining solutions by leveraging both static and dynamic information, until convergence is achieved or a predefined iteration limit is reached.

\subsection{Automatic Evaluation}

To enable scalable and cost-effective evaluation, we developed an automated system to assess the answers by verifying the satisfaction of each NPC's preferences.

For each problem, we check each NPC in the answer. Their embodied preferences are evaluated by the simulator. It calculates the assigned chair's position and distance from furniture, windows, etc., to determine whether embodied preferences are met. For example, if an NPC wants to watch television, the simulator checks if the television is within their field of view from that seat. For social preferences, a hand-crafted discriminator checks if the people beside the NPC meet their social preferences. For conflicts, we consider that the conflicts are avoided if the NPC's immediate neighbors do not have conflicts with them.

We calculated a weighted score for each question, as shown in Equation~\ref{eq:score}.
\begin{equation}
score_{k}=\sum_{c=1}^{k_c}w_c*\mathcal{F}_{c}(\frac{\sum_{i=1}^{n}\sum_{j=1}^{m_{i,c}}w_{i,j|c}*grade_{i,j|c}}{\sum_{i=1}^{n}\sum_{j=1}^{m_{i,c}}w_{i,j|c} }),
    \label{eq:score}
\end{equation}
 where $grade_{i,j|c}=1$ if this preference is satisfied, otherwise $grade_{i,j|c}=0$.

The $k$-th question involves $n$ NPCs, where each NPC $i$ has $m_{i,c}$ preferences that belongs to the $c$-th preference category. $grade_{i,j|c}$ indicates whether the $j$-th preference of NPC $i$ that belongs to $c$-th category is satisfied, and $w_{i,j|c}$ represents its weight. The weight is defined by the 3-point Likert strength upon generation. $w_c$ denotes the weight of the $c$-th category, defined as the sum of the weights of all preferences within that category.

The categories and preferences they contain are shown in Figure~\ref{fig:element}. To avoid a simple binary (0 or 1) evaluation, we compute the answer score at the category level. While an extreme view grants full credit only if all intra-group preferences are met (otherwise zero), this requirement is too rigid and lacks necessary granularity. Consequently, we employ a polynomial formula, $ \mathcal{F}(\cdot)$, based on empirical observation to remap the per-group scores to a continuous range of $[0, 1]$, as shown in Equation $\ref{eq:remap}$.  It places a higher penalty on answers that only satisfy a small fraction of preferences within a category.
\begin{equation}
    \mathcal{F}(x)=-10.87x^5+21.99x^4-12.65x^3+2.568x^2-0.045x
    \label{eq:remap}
\end{equation}

To analyze the T-Agent's capability in prioritizing based on weights, we statistically tracked its preference satisfaction based on their strengths. We define the prioritization gap (PG) as the satisfying rate of high-weighted preferences minus that of the low-weighted preferences.

To facilitate human review of the T-Agent’s performance, we scale the evaluation scores to a 0-100 range for enhanced clarity and offer several in-simulator visualizations of the T-Agent’s answers. These include: 1) Rendering the NPCs at their assigned seats according to the T-Agent's answer; 2) Displaying each NPC’s preferences above their head for easy comparison by human reviewers; 3) Drawing lines connecting NPCs to represent their relationships; and 4) Visualizing the summary of relationships and preferences as perceived by the T-Agent.

%% file: S3IT/5_Experiment.tex
\section{Experiment}
\subsection{Experimental Setup}
To ensure a comprehensive and reproducible evaluation, we constructed a standardized 70-question Set, hereafter the ``Test Set", from our large-scale S$^3$IT dataset. This subset was constructed by sampling one representative question from each of the 70 unique difficulty levels, a strategy that preserves the feature distribution of the full 7,000-question dataset (Figure \ref{fig:dataset}). The evaluation then proceeds through the three phases outlined in the Testing Pipeline section.
\begin{figure}[tbp]
\renewcommand{\baselinestretch}{1.0}
\centering 
\includegraphics[width=0.95\linewidth, keepaspectratio] {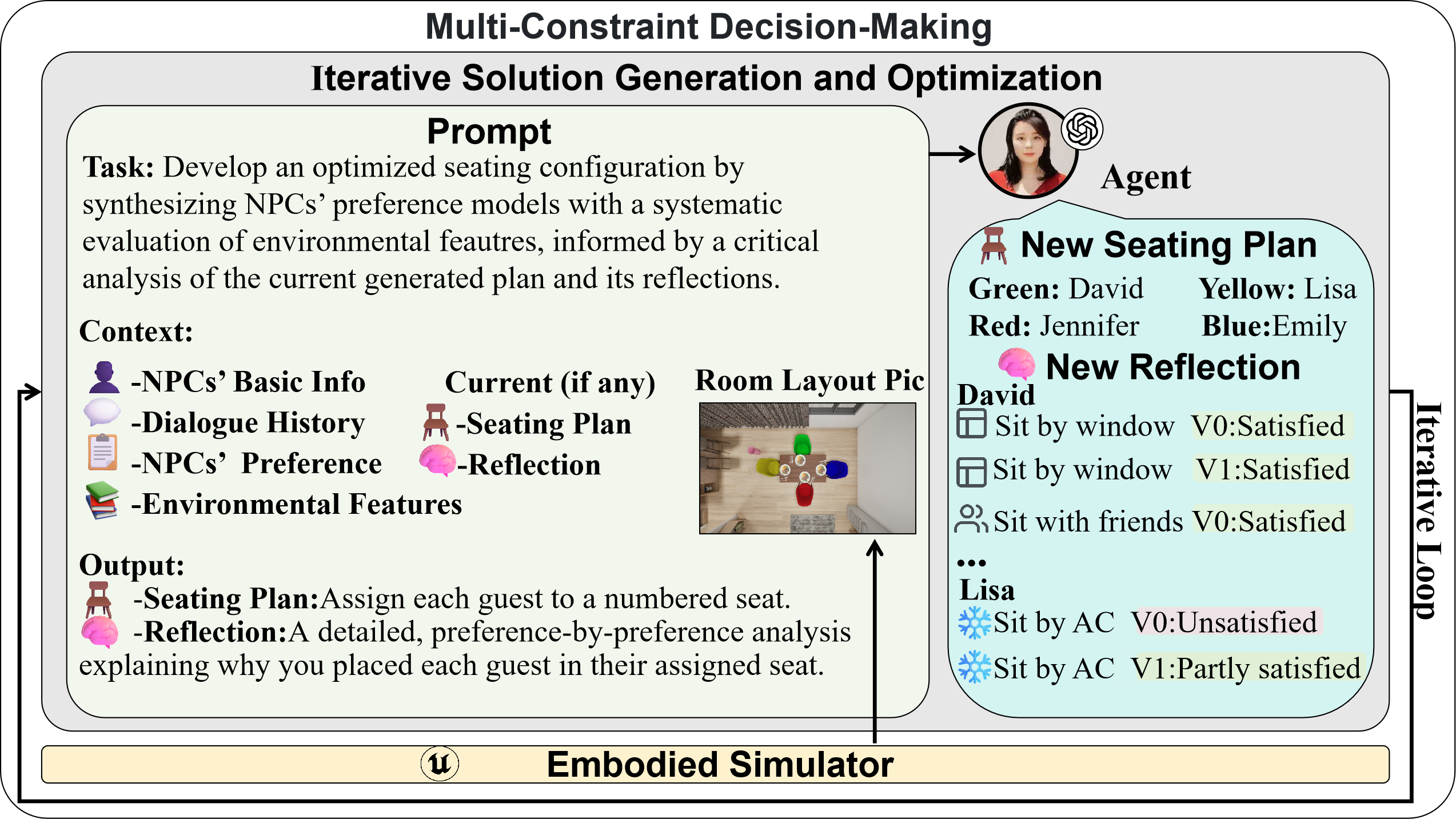} \vspace{-10pt}
\caption{
The T-Agent employs a ``generate-and-reflect'' mechanism to iteratively refine the solution. It integrates static constraints (NPCs' profiles, environmental features) with dynamic feedback from the preceding iteration (the generated plan and its reflection). Through this process, the T-Agent iteratively refines its approach, converging on a solution that satisfies multiple, complex constraints. Abbreviations: V0: Initial seating plan; V1: Optimized seating plan; AC: Air conditioner.} \label{phase3} \vspace{-6pt}
\end{figure}
\begin{table}[tb]
\centering
\caption{Model performances on the Test Set. Best results are in bold.} \vspace{4pt}
\label{tab:results_compact}
\renewcommand{\arraystretch}{0.9}
\setlength{\tabcolsep}{2pt}
\small
\begin{tabular}{lccccc}
\toprule
\textbf{Model} & \textbf{Embodied} & \textbf{Social}  & \textbf{Conflict} &\textbf{PG} & \textbf{Average} \\

\midrule
GPT-4o-mini & 16.7 & 34.7 & 45.8 &1.6 & 19.3 \\
Claude-4.5 & 19.1 & 37.6 & 46.0 &4.8 & 23.1 \\
GPT-4.1-mini & 22.8 & 39.3 & 42.5 &3.7 & 26.8 \\
GPT-4o & 24.6 & 43.0 & 51.7 &3.3 & 28.2 \\
Doubao-1.5 & 24.6 & 43.0 & 62.5 &3.8 & 28.3 \\
GPT-4.1 & 23.3 & 43.2 & 55.4 &3.8 & 29.3 \\
o4-mini & 29.0 & 54.5 & \textbf{89.5} &6.8 & 41.4 \\
GPT-5 & 29.0 & \textbf{56.9} & 86.1 &\textbf{15.4} & 42.7 \\
o3 & 32.9 & 53.8 & 89.0 &12.7 & 43.1 \\
Gemini-2.5-pro & \textbf{40.6} & 56.2 & 85.7 & 8.8& \textbf{47.8} \\
\bottomrule
\end{tabular}
\end{table}

We benchmarked a suite of state-of-the-art (SOTA) LLMs on our S$^3$IT benchmark, including OpenAI's GPT-5, GPT-4.1, GPT-4o, o4-mini, GPT-4o-mini, o3; Google's Gemini-2.5-Pro; Anthropic's Claude-4.5; and ByteDance's Doubao-1.5 (see Appendix for full model identifiers). 

For the human performance baseline, we recruited three participants. Since the full 70-question set was prohibitively long for human evaluation (est\(>\) 20 hours/person), we sampled a 10-question representative subset using an Integer Programming algorithm, hereafter referred to as the ``Human Test Set''. This set preserved the key feature distribution (e.g., preference types, conflict density, scene complexity) of the full set to ensure a fair comparison.

\begin{table}[tb]
\centering
\caption{Detailed performance on social preferences on the Test Set. Best results are in bold.} \vspace{4pt}
\label{tab:social_pref}
\renewcommand{\arraystretch}{0.9}
\setlength{\tabcolsep}{8pt}
\small
\begin{tabular}{lccc}
\toprule
\textbf{Model} & \textbf{Relation} & \textbf{Group of NPC} & \textbf{Topic} \\
\midrule

GPT-4o-mini & 37.4 & 34.7 & 45.7 \\
Claude-4.5 & 44.5 & 38.1 & 59.0 \\
GPT-4.1-mini & 44.1 & 41.6 & 52.7 \\
GPT-4o & 48.7 & 44.2 & 61.2 \\
Doubao-1.5 & 50.4 & 40.7 & 64.7 \\
GPT-4.1 & 63.9 & 43.3 & 54.0 \\
o4-mini & 66.2 & 51.2 & 64.9 \\
GPT-5 & 76.1 & 51.1 & \textbf{70.7} \\
o3 & 69.9 & 50.1 & 65.6 \\
Gemini-2.5-pro & \textbf{78.7} & \textbf{57.8} & 63.7 \\
\bottomrule
\end{tabular}
\end{table}

\begin{table}[!b]
\centering
\caption{Comparison of Human and SOTA LLM Performance on the Human Test Set.}  \vspace{4pt}
\label{tab:results_human}
\setlength{\tabcolsep}{2pt}
\small
\begin{tabular}{lccccc}
\toprule
\textbf{Model} & \textbf{Embodied} & \textbf{Social}  & \textbf{Conflict} & \textbf{PG}  & \textbf{Average} \\
\midrule
Gemini-2.5-pro & 36.3 & 51.0 & 89.3& 14.2& 41.4 \\
Human &90.6  & 77.9 & 89.8 &1.7 & 84.7\\
\bottomrule
\end{tabular}
\end{table}

\subsection{Results and Discussion}
The results in Table \ref{tab:results_compact} show that Gemini-2.5-pro emerged as the SOTA model with a top score of 47.8, and was notably the only model to exceed 40 on the challenging ``Embodied Preference" dimension. We attribute its leading performance to its strong native multimodal integration. The core challenge of S$^3$IT lies in effectively fusing dynamic spatial information from visual exploration with preferences from textual dialogue. Models with robust multimodal architectures can better establish mappings between cross-modal representations, granting them a decisive advantage in ``grounding" abstract preferences into concrete physical scenarios. Conversely, models like Claude-4.5 exhibit weaker performance, likely due to their optimization for different objectives, such as code completion. 
We also find that all models yield positive PG scores that scale with their capability, indicating an ability to distinguish between stronger and weaker preferences.

Our comparison with the human baseline (Table \ref{tab:results_human}) reveals that models significantly underperform relative to humans. Humans achieved an average score of 84.7, far surpassing the top-performing LLM's 41.4. The performance disparity is most pronounced on the ``Embodied" dimension, where humans scored 90.6 against Gemini-2.5-pro's 36.3. This highlights the critical shortcomings of current large models in spatial intelligence. In contrast, we found that top models achieve near-human performance on the ``Conflict" dimension, demonstrating their proficiency with explicit, rule-based constraints. Notably, the lower PG score for humans is attributed to a strategy of accommodating all preferences to maximize their overall score.

A fine-grained analysis of the scores (Table \ref{tab:social_pref}) indicates that models systematically underperform on ``Group of NPC" preferences compared to ``Relation" and ``Topic" categories. We attribute this disparity to different reasoning complexities. ``Relation" and ``Topic" preferences primarily require matching explicit cues in the context. In contrast, ``Group of NPC" preferences demand a multi-step, programmatic reasoning process: the agent must perform a global scan to extract NPC attributes, ground concepts like ``peers" or ``highly-educated group" into operational criteria, and then filter the candidate set. 
This performance gap highlights a limitation: models excel at information matching but struggle with tasks requiring multi-step reasoning.

\begin{table}[!b]
\centering
\caption{Impact of Ground-Truth Perception on Model Performance on the Test Set. An asterisk (*) denotes models evaluated with Ground-Truth (GT) perception.}
\vspace{4pt}
\label{tab:ablation_gt_updated}
\setlength{\tabcolsep}{2pt}
\small
\renewcommand{\arraystretch}{0.9}
\begin{tabular}{lccccc}
\toprule
\textbf{Model} & \textbf{Embodied} & \textbf{Social}  & \textbf{Conflict} &\textbf{PG}& \textbf{Average} \\
\midrule
Doubao-1.5         & 24.6 & 43.0 & 62.5&3.8 & 28.3 \\
Doubao-1.5* & 86.1 & 61.2 & 87.1 &11.1& 75.5 \\
\midrule 
o3                     & 32.9 & 53.8 & 89.0 &12.7& 43.1 \\
o3*             & 80.1 & 75.0 & 94.4 & 13.7&79.8 \\
\midrule 
Gemini-2.5-pro         & 40.6 & 56.2 & 85.7 &8.8& 47.8 \\
Gemini-2.5-pro* & 78.5 & 66.7 & 91.4 & 16.8&74.8 \\

\bottomrule
\end{tabular}
\end{table}

\subsection{Ablation Study}
We conducted an oracle ablation study to test our hypothesis that spatial intelligence is the key performance bottleneck. By directly providing ground-truth (GT) perception, we simulated perfect spatial awareness. As detailed in Table \ref{tab:ablation_gt_updated}, this led to dramatic score increases across all tested models.
These findings confirm that the primary limitation is not reasoning but the inability to derive structured 3D understanding from visual inputs. The concurrent improvement in ``Social" and ``Conflict" scores further validates our benchmark's reliance on spatial cues, proving that even social tasks in our setting require more than just textual inference. Notably, when equipped with GT perception, the models' performance approaches human levels (Table \ref{tab:results_human}), underscoring the latent reasoning potential of large models.

\begin{figure}
    \centering
    \includegraphics[width=0.9\linewidth]{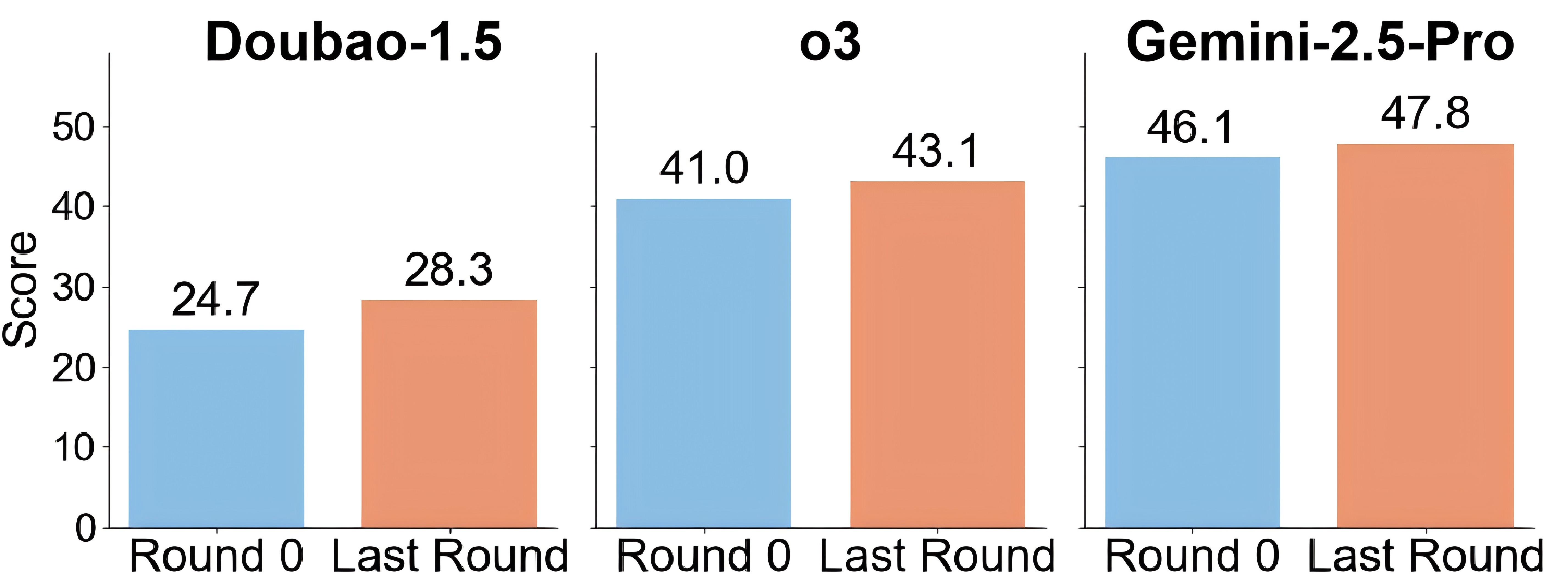}\vspace{-4pt}
    \caption{Model scores (The average score in Table~\ref{tab:results_compact}) in the initial round and after reflection.}
    \label{fig:reflection} \vspace{-12pt}
\end{figure}

Finally, we evaluated the efficacy of a self-reflection mechanism for error reduction. As illustrated in Figure \ref{fig:reflection}, which shows the change in average scores for three representative models across reflection iterations, all tested models demonstrated consistent improvement.
These results confirm the efficacy of our proposed ``generate-and-reflect" framework. Furthermore, in comparing this to human problem-solving strategies, we observed that human participants often adopt a heuristic-based ``anchoring" method. They tend to establish an initial, seemingly viable arrangement and, if this path proves infeasible, prefer making local adjustments over systematic backtracking. Consequently, they are often trapped by their initial, flawed assumptions. In contrast, the agent's computational persistence and systematic exploration allow it to mitigate such cognitive biases, including cognitive fatigue and heuristic shortcuts. It is noteworthy that the human participants required an average of 165 minutes to complete the 10-problem test suite, which underscores the substantial complexity of the tasks.

%% file: S3IT/6_Discussion.tex
\section{Limitation and Future Work}

Unlike narrative-based testing, S$^3$IT is interactive, where varying interaction processes can influence the performance. We've provided a typical interaction process for integration of T-Agents, which can also be extended to more complex forms for richer test complexity, such as removing the discrete viewpoint set and requiring the T-Agent to plan continuous motion trajectories, as well as configuring NPCs to behave uncooperatively when queried about their preferences.
Furthermore, our human evaluation relied on a small sample of three experts, which limits the generalizability of this performance baseline to a broader population.

We adopted an empirical curve to penalize models for satisfying only a small fraction of preferences; this score may include subjective bias. Critically, the curve remains consistent across all categories, ensuring it does not compromise the rank order of scores between LLMs and the human baseline.

%% file: S3IT/7_Conclusion.tex
\section{Conclusion}
We introduce S$^3$IT, a benchmark for embodied social intelligence featuring procedural generation and a multi-phase evaluation pipeline. Our evaluation of state-of-the-art LLMs reveals their proficiency in explicit rule-following but a critical deficit in integrating spatial and social reasoning, resulting in a stark performance gap against the human baseline. Our findings underscore that grounding abstract social cognition in physical environments is a key challenge for future research. We posit S$^3$IT will catalyze progress towards truly collaborative embodied agents.